\documentclass[a4paper,11pt]{article}

\usepackage{jinstpub_1} 

\title{\boldmath Improvement of training set structure in fusion data cleaning using Time-Domain Global Similarity method}

\author[a]{Jian. Liu,}
\author[a, 1]{Ting. Lan,\note{Corresponding author.}}
\author[a, b]{Hong. Qin,}

\affiliation[a]{School of Nuclear Science and Technology and Department of Modern Physics, University of Science and Technology of China, Hefei, Anhui 230026, China}
\affiliation[b]{Plasma Physics Laboratory, Princeton University, Princeton, NJ 08543, USA}

\emailAdd{lanting@ustc.edu.cn}

\abstract{Traditional data cleaning identifies dirty data by classifying original data sequences, which is a class$-$imbalanced problem since the proportion of incorrect data is much less than the proportion of correct ones for most diagnostic systems in Magnetic Confinement Fusion (MCF) devices. When using machine learning algorithms to classify diagnostic data based on class$-$imbalanced training set, most classifiers are biased towards the major class and show very poor classification rates on the minor class. By transforming the direct classification problem about original data sequences into a classification problem about the physical similarity between data sequences, the class$-$balanced effect of Time$-$Domain Global Similarity (TDGS) method on training set structure is investigated in this paper. Meanwhile, the impact of improved training set structure on data cleaning performance of TDGS method is demonstrated with an application example in EAST POlarimetry$-$INTerferometry (POINT) system.}

\begin{document}
\maketitle
\flushbottom

\section{Introduction}
\label{sec:intro}

To analyze diagnostic data in fusion experiments effectively, it is necessary to seek an automatic data cleaning method, which can sort out incorrect data from massive original diagnostic data accurately and quickly. In traditional opinion, data cleaning can be treated as a typical binary classification problem, i.e., how to properly divide the original data set into two groups, correct data sequences and incorrect ones. For most diagnostic systems in Magnetic Confinement Fusion (MCF) devices, the proportion of incorrect diagnostic data is much less than the proportion of correct ones. That means the class structure of database  is imbalanced. When using machine learning algorithms to classify diagnostic data based on class$-$imbalanced training set, most classifiers are biased towards the major class and show very poor classification rates on the minor class~\cite{japkowicz2002class,wei2013role}. So the lack of dirty data in original diagnostic database leads to poor data cleaning performance by using traditional classification algorithms directly.

Recently, a new data cleaning method, called Time-domain Global Similarity (TDGS) method  \cite{2017arXiv170504947L}, has been proposed. The TDGS method is a general-purposed classification method based on machine learning techniques, which can be used to classify the original diagnostic data into a correct and an incorrect group. Unlike traditional classification methods in machine learning, TDGS method focuses on the classification of physical similarity between diagnostic data sequences, instead of the direct classification of original data itself. This new idea enables TDGS method much wider application prospects, because physical similarity reflects intrinsic physical relevance between data sequences from different measuring channels. Traditional data sorting aims to the classification of original diagnostic data sequences. The corresponding class structure is reflected by $\ensuremath{R_J}$, i.e., the ratio of incorrect data to correct ones. The focus of TDGS method turns to the physical similarity between diagnostic data sequences. The class structure of training set in TDGS method is depicted by $\ensuremath{R_{TDGS}}$, i.e., the ratio of dissimilar samples to similar ones. By transforming the direct classification problem about original data sequences into a classification problem about the physical similarity between data sequences, the structure of training set can be improved by TDGS method. 

In this paper, the class$-$balanced effect of TDGS method on the structure of training set is investigated. Meanwhile, the impact of improved training set structure on data cleaning performance of TDGS method is demonstrated with an application example in EAST POlarimetry$-$INTerferometry (POINT) system. Each sample of TDGS method is generated by combining two data sequences from different channels of MUlti-channel Measurement (MUM) system. Most diagnostic systems of MCF
devices are MUM systems, which measure related yet distinct aspects
of the same observed object with multiple independent measuring channels,
such as common interferometer systems \cite{kawahata1999far}, polarimeter
systems \cite{donne2004poloidal,brower2001multichannel,liu2014faraday,liu2016internal,zou2016optical},
electron cyclotron emission imaging systems \cite{luo2014quasi},
etc.  From the diagnostic data of an N-channel MUM system for
P discharges, $\ensuremath{P*C_{N}^{2}}$ samples can be generated.
And sample tag is set as the corresponding physical similarity between these two sequences. By tagging the sample consist by two correct data sequences as similarity,
$\ensuremath{\sum\limits _{i=1}^{P}{C_{N(1-{Q_{i}})}^{2}}}$ similar
samples can be generated, where $\ensuremath{{Q_{i}}}$ is the ratio
of incorrect data sequences to total data sequences for the $\ensuremath{ith}$
discharge. And $\ensuremath{P*C_{N}^{2}{\rm {-}}\sum\limits _{i=1}^{P}{C_{N(1-{Q_{i}})}^{2}}}$
dissimilar samples can be generated, which contain at least one incorrect
data sequence. By selecting the parameters $\ensuremath{N}$ and $\ensuremath{{Q_{i}}}$, the class structure of training set can be balanced. By comparing the performance of classifiers generated from training sets of various class structures, the impact of improved training set structure on fusion data cleaning can be exhibited. 

The rest parts of this paper are organized as follows. In section~\ref{sec:2}, the class$-$balanced effect of TDGS method on training set structure is explained. In section~\ref{sec:3}, as an example, the impact of improved training set structure on data cleaning performance of TDGS method is demonstrated with applications in density data of various class structures from POINT system. In section~\ref{sec:4}, the further improvements of applying TDGS method to class-imbalanced database are discussed.

\section{Class-balanced effect of TDGS method on training set structure}
\label{sec:2}

The class structure defined in TDGS method is the ratio of
dissimilar samples to similar ones, which is different from the class
structure defined in traditional data cleaning, i.e., the ratio of
incorrect data sequences to correct ones. In this section, the balanced effect of TDGS method on the class structure
of training set is explained. 

\begin{figure}[htbp]
\centering 
\includegraphics[scale=0.4]{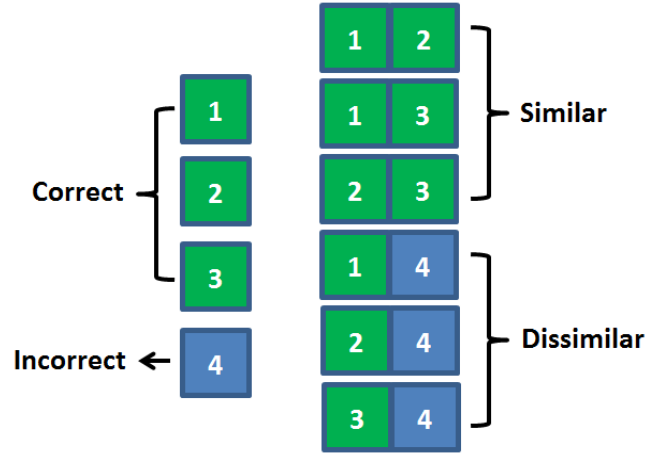}
\caption{\label{fig:1} The class structure transformation of TDGS method is shown with an
example in the database from a 4-channel MUM system for one discharge.}
\end{figure}

The transformation of TDGS
method on class structure can be exhibited with the database from a 4-channel
MUM system for one discharge, see figure~\ref{fig:1}. In this example,
the ratio of incorrect data to correct ones is $\left. {1} \middle/ {3} \right.$,
which is the class structure in directly classifying the original
data sequences. By combining two data sequences from different
channels as one sample, 6 samples are generated by TDGS method. Among them, 3
samples constituted by two correct data sequences are tagged with
similarity, and the other 3 samples containing at least one incorrect
data sequence are tagged with dissimilarity. The class structure of
TDGS method is $\left. {1} \middle/ {1} \right.$. After the transformation
of TDGS method, the class structure of training set is more balanced in this case.

\begin{figure}[htbp]
\centering 
\includegraphics[scale=0.3]{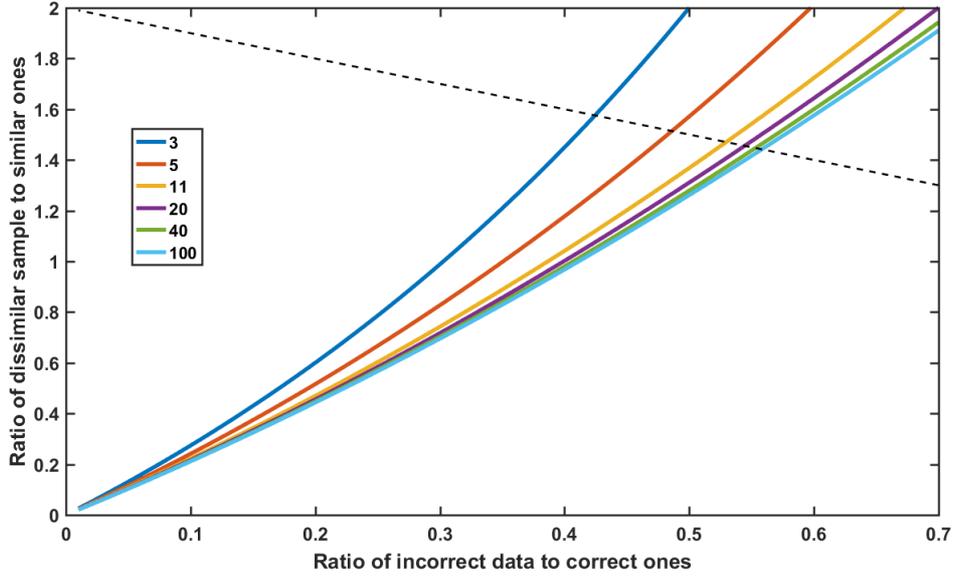}
\caption{\label{fig:2} When the ratio of incorrect data sequences to total data sequences
for each discharge is equal, the class structure transformation curves
of TDGS method for some common MUM systems are plotted. The legend
number denotes corresponding channel numbers of MUM systems. The region
below the black dashed line is the class-balanced area of TDGS method.}
\end{figure}

From the diagnostic data of an N-channel MUM system for P
discharges, $\ensuremath{P*C_{N}^{2}}$ samples can be generated by
combining two data sequences from different channels under the
same discharge. Suppose the correct data sequences for the $\ensuremath{ith}$
discharge are $\ensuremath{N(1-{Q_{i}})}$ , where $\ensuremath{{Q_{i}}}$
is the ratio of incorrect data sequences to total data sequences for
corresponding discharge. By combining two correct data sequences, $\ensuremath{C_{N(1-{Q_{i}})}^{2}}$ similar
samples can be generated for the $\ensuremath{ith}$ discharge. And $\ensuremath{\sum\limits _{i=1}^{P}{C_{N(1-{Q_{i}})}^{2}}}$
similar samples can be generated for P discharges. Apart from similar
samples, the other part is dissimilar samples, i.e., $\ensuremath{P*C_{N}^{2}{\rm {-}}\sum\limits _{i=1}^{P}{C_{N(1-{Q_{i}})}^{2}}}$
dissimilar samples can be generated. In this general case, the class structure
of TDGS method is $\left. {\{ P*C_N^2 - \sum\limits_{i = 1}^P {C_{N(1 - {Q_i})}^2} \} } \middle/ {\sum\limits_{i = 1}^P {C_{N(1 - {Q_i})}^2} } \right.$. When the proportion of incorrect data sequences for each discharge is equal, the class structure transformation curves
of TDGS method for some common MUM systems are plotted in figure~\,\ref{fig:2}.
The region below the black dashed line is the class-balanced area
of TDGS method, where the class structure of TDGS method is more balanced
than the class structure of original data sequences, i.e.,
\begin{equation}
\label{eq:3}
\begin{split}
\ensuremath{\left|{{{Dissimilar}\mathord{\left/{\vphantom{{Dissimilar}{Similar}}}\right.\kern -\nulldelimiterspace}{Similar}}-1}\right|\le\left|{1-{{Incorrect}\mathord{\left/{\vphantom{{Incorrect}{Correct}}}\right.\kern -\nulldelimiterspace}{Correct}}}\right|.}
\end{split}
\end{equation}

The intersection range of class-balanced area and the class structure
transformation curve is wider for MUM system with more channels,
which indicates that TDGS method has better balanced effect for MUM
system with more channels. When the ratio of incorrect data sequences
to correct ones for each discharge is equal and below 0.4, TDGS method
has balanced effect for common MUM systems, see figure~\ref{fig:2}.

\section{Applications of TDGS method in class-imbalanced density data from POINT system}
\label{sec:3}

In this section, the performance of TDGS method on class-imbalanced
data is shown with an application example in cleaning density data from POINT system. By comparing the performance of classifiers generated
from training sets of various class structures, the impact of training set structure on TDGS method is exhibited. 

\begin{figure}[htbp]
\centering 
\includegraphics[scale=0.3]{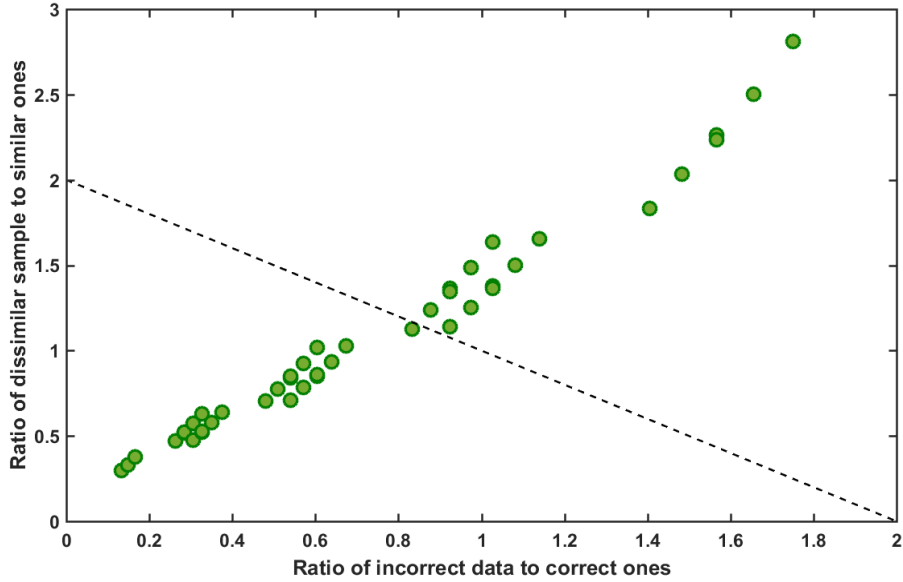}
\caption{\label{fig:3} For the selected training sets from POINT system, the class structure
transformation curves of TDGS method are plotted. The region below
the black dashed line is the class-balanced area of TDGS method.}
\end{figure}

POINT is a typical 11-channel MUM system, which measures
line-average electron density of EAST tokamak at different vertical
locations with independent measuring channels \cite{liu2014faraday,liu2016internal,zou2016optical}.
In this application, density data of POINT system for 7 discharges
are chosen as training set. By combining two data sequences from different
channels, $\ensuremath{7*C_{11}^{2}=385}$ samples
are generated. To compare the performance of classifiers
generated from training sets of different class structures, the training
set are selected from data for 12 discharges of various error rates,
i.e., $\ensuremath{C_{12}^{7}=792}$ training sets of multiple class
structures are generated. Here the error rate for each discharge denotes
the ratio of incorrect data sequences to total data sequences.
For the selected 792 training sets, the class structure transformation
curves of TDGS method are plotted in figure~\ref{fig:3}. The region
below the black dashed line is the class-balanced area of TDGS method.
In conventional operations of POINT system, the mean ratio of incorrect
data to correct ones is much less than 1, which is involved in the
class-balanced area.

In the training process of this application, Support Vector
Machine (SVM) is adopted as the classification algorithm for the advantage
in solving non-linear, high-dimensional problems \cite{cortes1995support,boser1992training,hsu2003practical}.
In SVM, input samples are mapped to a high-dimensional feature space.
A good classification is achieved by constructing a linear separating
hyperplane in this feature space with the maximal margin to the nearest
samples of any class. Here sequential minimal optimization (SMO) is
adopted as the iterative method for solving this quadratic programming
(QP) problem \cite{platt1998sequential}. Proper selection of kernel
function for corresponding classification problem can optimize the
performance by mapping samples to appropriate feature space. In this
application, linear kernel function is chosen for it has less kernel
parameters to be optimized and faster training speed \cite{hsu2003practical}.
Meanwhile, the penalty parameter of the error term is set to 20. After
training in the dataset of various class structures, different classifiers
for data cleaning can be generated.

The performance of classifiers generated from training set
of various class structures is assessed in the same validation set.
Here density data from other 12 discharges of low error rates are
selected as validation set, which is consistent with the real data
characteristics of POINT system in conventional operations. Training
sets of an identical class structure can be categorized as the same
group. To provide an unbiased error estimate, the performance of TDGS
method on corresponding class structure is estimated by taking the
average results in training sets of the same group. Meanwhile, the
geometric mean (G-mean) of recall rates observed separately on positive
examples and negative examples is a common assessment measure for
class-imbalanced problem \cite{kubat1997addressing}, which is defined
as
\begin{equation}
\label{eq:4}
\begin{split}
\ensuremath{{\{[TP/(TP+FN)]*[TN/(TN+FP)]\}^{1/2}}.}
\end{split}
\end{equation}

\begin{figure}[htbp]
\centering 
\includegraphics[scale=0.3]{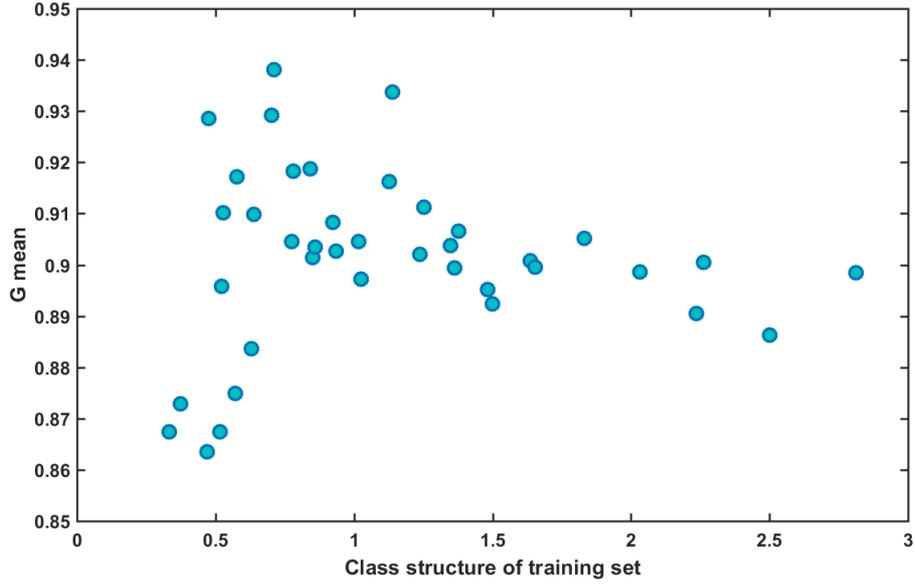}
\caption{\label{fig:4} The assessment results of applying classifiers generated from training
set of various class structures to the same validation set. The changes
of assessment measure G-mean versus class structure of training set
are plotted. Class structure of training set denotes the ratio of dissimilar
samples to similar ones.}
\end{figure}

In this case, TP is the number of dissimilar samples which
are correctly classified, FN is the dissimilar samples which are incorrectly
classified as similar ones, FP is the similar samples which are incorrectly
classified as dissimilar ones, and TN is the similar samples which
are correctly classified. The assessment results of applying classifiers
generated from training set of various class structures to the same
validation set are shown in figure~\ref{fig:4}. It can be observed
that performance of classifiers is better when the class structure
of training set is more balanced, i.e., the ratio of dissimilar samples
to similar ones is closer to 1. Then a good data cleaning performance
can be achieved by training with a class-balanced training set with
TDGS method.

\section{Summary}
\label{sec:4}

Machine learning has advantages in cleaning fusion data for
MCF science. Choosing a class-balanced training set favors to generate
efficient classifiers for data cleaning. While the class structures
of original diagnostic fusion data are not balanced. We have proposed
TDGS method to automatically sort out dirty diagnostic data of MUM
systems, which has a new definition about the class structure. In
this paper, the balanced effect of TDGS method on the class structure
of database is investigated. By selecting the parameters $\ensuremath{N}$
and $\ensuremath{{Q_{i}}}$, the class structure of training set can be balanced. Meanwhile, the performance of applying TDGS
method to class-imbalanced data is demonstrated with an application
example in database of various class structures. The assessment results
show that a class-balanced training set is beneficial for achieving
a good data cleaning performance with TDGS method.

There exits some common techniques for handling class-imbalanced database in machine learning, such as under-sampling, over-sampling, and cost-modifying. Next step, we would compare these techniques with TDGS method in solving class-imbalanced problems.
Based on corresponding characteristics and application ranges of these methods, a proper combination of traditional techniques with TDGS method would further improve the data cleaning performance for class-imbalanced data.  

\acknowledgments

This research is supported by Key Research Program of Frontier Sciences CAS (QYZDB-SSW-SYS004), National Natural Science Foundation of China
(NSFC-11575185,11575186), National Magnetic Confinement Fusion Energy
Research Project (2015GB111003,2014GB124005), JSPS-NRF-NSFC A3 Foresight
Program (NSFC-11261140328),and the GeoAlgorithmic Plasma Simulator
(GAPS) Project. 


\bibliographystyle{apsrev}
\bibliography{reference}

\end{document}